\documentclass[conference]{IEEEtran}
\IEEEoverridecommandlockouts
\usepackage{cite}
\usepackage{amsmath,amssymb,amsfonts}
\usepackage{algorithmic}
\usepackage{graphicx}
\usepackage{textcomp}
\usepackage{xcolor}
\usepackage{dblfloatfix}
\usepackage{amsfonts}
\usepackage{dsfont}

\usepackage{bbm} 
\usepackage{subfigure}
\usepackage{amsmath,amsthm,bm}
\usepackage{amssymb}
\usepackage{flushend}
\usepackage{makecell}

\def\BibTeX{{\rm B\kern-.05em{\sc i\kern-.025em b}\kern-.08em
    T\kern-.1667em\lower.7ex\hbox{E}\kern-.125emX}}
\begin{document}

\title{Self-Contrastive Learning based Semi-Supervised Radio Modulation Classification
}

\author{\IEEEauthorblockN{Dongxin Liu\IEEEauthorrefmark{1},
Peng Wang\IEEEauthorrefmark{2},
Tianshi Wang\IEEEauthorrefmark{1},
and
Tarek Abdelzaher\IEEEauthorrefmark{1}}
\IEEEauthorblockA{\IEEEauthorrefmark{1}Department of Computer Science,
University of Illinois Urbana-Champaign, Champaign, IL, 61820, USA}
\IEEEauthorblockA{\IEEEauthorrefmark{2}US Army Research Laboratory, Aberdeen Proving Ground, MD, 21005, USA}
Email: \{dongxin3, tianshi3, zaher\}@illinois.edu; peng.wang2.civ@mail.mil
}
\maketitle

\begin{abstract}
This paper presents a semi-supervised learning framework that is new in being designed for {\em automatic modulation classification\/} (AMC). By carefully utilizing {\em unlabeled\/} signal data with a self-supervised contrastive-learning pre-training step, our framework achieves higher performance given smaller amounts of {\em labeled\/} data, thereby largely reducing the labeling burden of deep learning. We evaluate the performance of our semi-supervised framework on a public dataset. The evaluation results demonstrate that our semi-supervised approach significantly outperforms supervised frameworks thereby substantially enhancing our ability to train deep neural networks for automatic modulation classification in a manner that leverages unlabeled data.
\end{abstract}

\begin{IEEEkeywords}
Self-Supervised Learning, Semi-Supervised Learning, Automatic Modulation Classification
\end{IEEEkeywords}

\section{Introduction}
Automatically recognizing or classifying the radio modulation is a key step for many commercial and military applications, such as dynamic spectrum access, radio fault detection, and unauthorized signal detection in battlefield scenarios. The modulation classification problem has gotten widely studied in the past few years. Two general types of algorithms, likelihood-based (LB) and feature-based (FB), have been applied to solve the modulation classification problem. Likelihood-based methods~\cite{long1994further,lay1995modulation} make decisions based on the likelihood of the radio signal and achieve optimal performance in the Bayesian sense. However, they suffer from high computational complexity. Feature-based methods~\cite{dobre2005classification,ho2000modulation,huang2016automatic}, on the other hand, make decision based on the manually-extracted features from the radio signal. Given the input radio signal, various features related to phase, amplitude or frequency are manually extracted and then used as inputs of the classification algorithms. Machine learning algorithms, such as support vector machine (SVM) and decision trees, are popular candidates for the classification algorithms.

Recently, with advances in deep learning techniques, deep neural networks achieves a great success in multiple fields, such as image classification~\cite{krizhevsky2012imagenet} and natural language processing~\cite{devlin2018bert}. Deep neural network models, such as convolutional neural networks (CNNs)~\cite{o2016convolutional, o2018over} and recurrent neural networks (RNNs)~\cite{rajendran2018deep, xu2020spatiotemporal}, have also been applied to the modulation classification problem. Such neural network models directly feed the raw signal data or its transforms as input and generally achieve much better performance than previous approaches. 

However, training a deep neural network model requires a large amount of training data. In practice, it's usually difficult to collect a large volume of high quality and reliable radio signals as well as their modulations (labels) as training data. One common way to deal with the lack of training data is data augmentation. Previous studies proposed several data augmentation strategies \cite{huang2019data,zheng2020spectrum,wang2019data} to avoid overfitting caused by the lack of training data. Another way is to utilize {\em unlabeled data\/}. The difficulty in collecting a large training dataset for radio modulation classification mainly comes from the labeling burden for radio signals. It is much eaiser to collect the radio signals without labeling them. Hence, it would help a lot if we could efficiently utilize {\em unlabeled\/} radio signals. As far as we know, the benefits of using unlabeled data have not yet been studied for the modulation classification problem. 

Self-supervised learning, as a kind of unsupervised learning, obviates much of the labeling burden by extracting information from unlabeled data. Self-supervised learning algorithms have had great success in areas of computer vision~\cite{chen2020simple}, natural language processing~\cite{devlin2018bert}, and IoT applications~\cite{liu2021contrastive}. The goal of self-supervised learning is to train an encoder to extract useful intrinsic information (or features) from unlabeled data and  use that information as the inputs to a classifier or predictor in a downstream task. Since these features already store a lot of intrinsic information about the original input, a simple classifier (e.g., linear) is enough for the downstream task. Using this approach, most training uses unlabled data to learn the instrinsic features. Only a small amount of labeled data is then needed to train the downstream classifier. 

In this paper, we build a \textit{Semi}-supervised \textit{A}utomatic \textit{M}odulation \textit{C}lassification framework, namely, \textit{SemiAMC}, to efficiently utilize unlabeled radio signals. SemiAMC consists of two parts: (i) self-supervised contrastive pre-training and a (ii) downstream classifier. In self-supervised contrastive pre-training, we apply the design of SimCLR \cite{chen2020simple}, which is an effective self-supervised contrastive learning framework, to train an encoder using a large amount of unlabeled training data as well as data augmentation. Then, we freeze the parameters of the encoder and train a classifier, which takes the representations (output of the encoder) as input,  based on a small amount of labeled training data. 

We evaluate the performance of SemiAMC on a widely-used modulation classification dataset RadioML2016.10a\cite{o2016radio}. The evaluation results demonstrate that SemiAMC efficiently utilizes unlabeled training data to improve classification accuracy. Compared with previous supervised neural network models, our approach achieves a better accuracy given the same number of labeled training samples.

The rest of this paper is organized as follows. We introduce related work in Section~\ref{sec:related_work}. Section~\ref{sec:approach} describes the detailed design and implementation of SemiAMC. We cover the evaluation results in Section~\ref{sec:evaluation} and finally summarize this paper in Section~\ref{sec:conclusion}.

\section{Related Work} \label{sec:related_work}
In this section, we briefly introduce related background on automatic modulation classification, self-supervised and semi-supervised learning techniques.

\subsection{Deep Learning in Automatic Modulation Classification}
Motivated by the success of deep learning techniques in computer vision \cite{krizhevsky2012imagenet} and NLP \cite{devlin2018bert}, deep neural network models, such as CNN \cite{o2016convolutional}, ResNet \cite{o2018over}, and LSTM \cite{rajendran2018deep} have been applied to the automatic modulation classification task. Such neural network models directly take radio signals as input and predict the type of modulation as output. To further improve the performance of automatic modulation recognition, specific characteristics of the modulated radio signals are considered. For example, Zeng et al. \cite{zeng2019spectrum} use spectrograms, generated from the radio signals through short-time discrete Fourier transform, as the input to a CNN classifier. Perenda et al. \cite{perenda2019automatic} separate the amplitude and phase series of the input and train based on a parallel fusion method. Most of the previous works focused on designing supervised models while we design a semi-supervised learning framework in this paper.

\subsection{Semi-Supervised Learning}
Semi-supervised learning is a learning paradigm that leverages both labeled and unlabeled data to train machine learning models. With the development of deep learning techniques, deep neural network models become deeper and deeper and we need more and more labeled data to train such models. However, the collection of large datasets is very costly and time-consuming as it requires a lot of human labor work to annotate the dataset. In order to reduce the data labeling burden, a lot of semi-supervised learning algorithms \cite{van2020survey} have been designed to make use of large number of unlabeled data as well as a small number of labeled data. In this paper, our semi-supervised approach is based on self-supervised contrastive learning, which has gotten great success in computer vision area. To our best knowledge, we are the first who apply self-supervised contrastive learning to automatic modulation classification problem.

\subsection{Self-Supervised Learning}
Self-supervised learning is a learning paradigm where the model is trained on unlabeled data. 
Self-supervised contrastive learning is an typical self-supervised learning algorithm and has achieved outstanding performance \cite{chen2020simple}. The supervision of self-supervised contrastive learning comes from the user-designed pretext tasks which generate locally distorted data samples. A loss function can thus be generated that prefers mapping such similar inputs to nearby locations in the latent space. 

Self-supervised learning can be used as the pre-training step in the semi-supervised learning algorithms \cite{chen2020simple,liu2021contrastive}. An encoder that collects intrinsic information from the original input (and maps it to a latent space) is trained with a self-supervised learning algorithm based on large amounts of unlabeled data. A small amount of labeled data can then be used to train a model on these features.

\section{Approach} \label{sec:approach}
In this part, we first introduce the signal model we consider in this paper. Then, we describe the design of SemiAMC. 

\subsection{Signal Model}
We consider a single-input single-output communication system where we sample the in-phase and quadrature components of a radio signal through an analog to digital converter. The received radio signal $r(t)$ can be represented as
\begin{equation}
    r(t)=c*s(t)+n(t),
\end{equation}
where $s(t)$ refers to the modulated signal from the transmitter, $c$ refers to the path loss or gain term on the signal, and $n(t)$ is the Gaussian noise. The received radio signal $r(t)$ is sampled $N$ times at some sampling rate to obtain a length $N$ vector of complex values. In this paper, we treat the complex valued input as a 2-dimensional real-valued input. We denote it as $X$, where $X$ is a $2\times N$ matrix containing the in-phase (I) and quadrature (Q) components of the $N$ received signal samples. 

The goal of this paper is to determine the modulation type for any given radio signal $X$.

\subsection{Overview of SemiAMC}
\begin{figure*}[t]
  \centering
  \includegraphics[width=150mm]{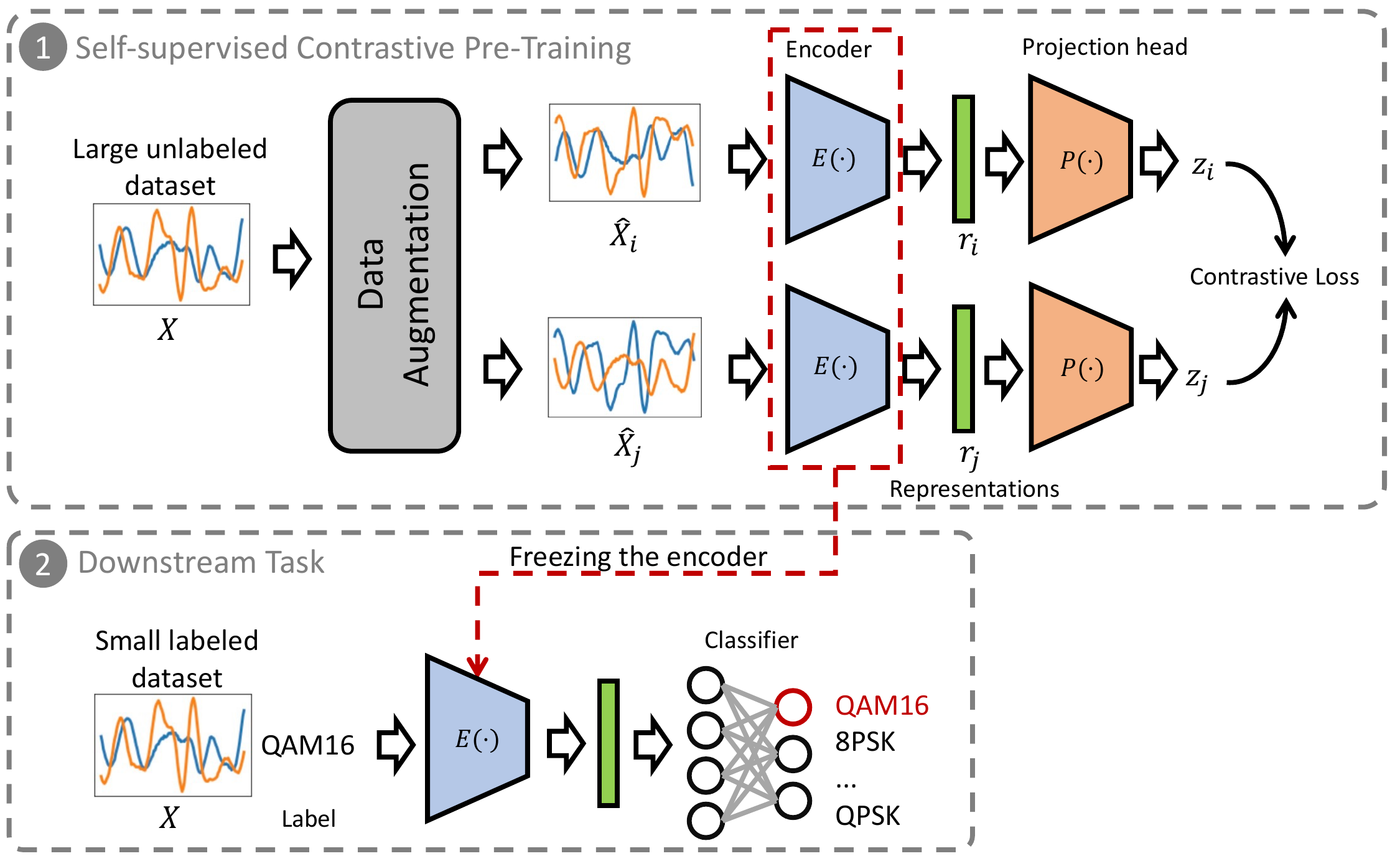}
  \caption{Overview of SemiAMC. We first train an encoder with a self-supervised contrastive learning framework using only unlabeled data. And then, the learned encoder and the corresponding features can be utilized to train the classifier with a small amount of labeled data.}
  \label{fig:overview}
\end{figure*}
SemiAMC aims at training a classifier to accurately recognize the modulation type for any given radio signal. As a semi-supervised framework, SemiAMC is trained with both labeled and unlabeled data. We show the architecture of SemiAMC in Figure~\ref{fig:overview}. The illustrated workflow is as follows. 

The first step is called self-supervised contrastive pre-training, where we train an encoder to map the original radio measurements into low-dimensional representations. This is done in a self-supervised manner, with unlabeled data only. The supervision here comes from optimizing the {\em contrastive loss\/} function, that maximizes the agreement between the representations of differently augmented views for the same data sample.

In step two, we freeze the encoder learned during the self-supervised contrastive pre-training step, and map the labeled input radio signals to their corresponding representations in the low-dimensional space. The classifier can be trained based on these representations and their corresponding labels. Here a relatively simple classifier (e.g., linear model) usually work well, because the latent representation has already extracted the intrinsic information from the signal input. In this way, a small number of labeled data samples is enough to train the classifier. When we have enough labeled data, we can also fine-tune the last one or more layers of the encoder to further improve the performance of SemiAMC.

\subsection{Self-supervised Contrastive Pre-training}
We apply SimCLR~\cite{chen2020simple} as our self-supervised contrastive pre-training framework. As shown in Figure~\ref{fig:overview}, our self-supervised contrastive pre-training mainly consists of four components.

\subsubsection{Data Augmentation}
The first component is a stochastic data augmentation module. Given any signal input $X$, two views of $X$, that are denoted as $\hat{X}_i$ and $\hat{X}_j$, are generated from the same family of data augmentation operations. The data augmentation algorithms are highly application dependent. For example, in image related applications, random color distortion, cropping, and Gaussian blue are commonly used data augmentation algorithms. In our approach, we augment the I/Q signal with the rotation operation \cite{huang2019data} which could keep the features for classification. For a modulated signal $X=[I,Q]^T$, where $I$ and $Q$ refer to $1\times N$ vectors storing the in-phase (I) and quadrature (Q) signals, we rotate it with an angle $\theta$ randomly selected from $\{0, \pi/2, \pi, 3\pi/2\}$. The augmented signal sample is 
\begin{equation}
    \hat{X} = \begin{bmatrix}\hat{I}\\\hat{Q}\end{bmatrix} = \begin{bmatrix} \cos\theta & -\sin\theta \\ \sin\theta & \cos\theta \end{bmatrix}\begin{bmatrix}I\\Q\end{bmatrix}.
\end{equation}

\subsubsection{Encoder}  
\begin{figure}[t]
  \centering
  \includegraphics[width=30mm]{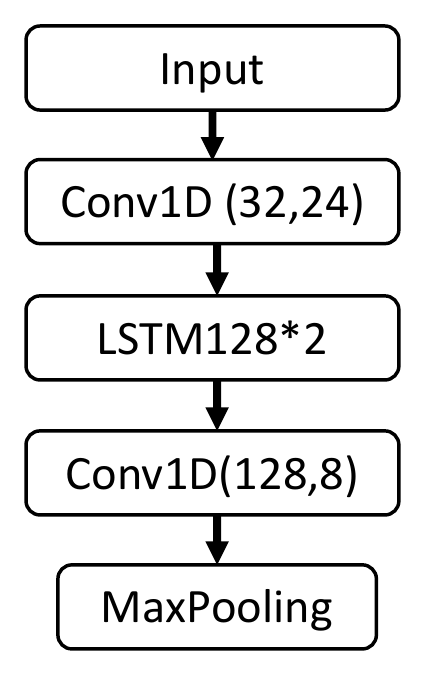}
  \caption{The architecture of the encoder.}
     \vspace{-4mm}
  \label{fig:encoder}
\end{figure}
The second part is a neural network based encoder $E$ that extracts intrinsic information from the augmented examples $\hat{X}_i$ and $\hat{X}_j$, and stores them in the latent representations $\bm{r}_i$ and $\bm{r}_j$:
\begin{equation}
    \bm{r}_i = E(\hat{X}_i),\ \bm{r}_j = E(\hat{X}_j).
\end{equation}
Various choices of the network architectures can be used to design the encoder. For example, CNN-based encoders are widely utilized in learning representations of visual data and RNN-based encoders are usually utilized to deal with time-series. Figure~\ref{fig:encoder} shows the architecture of the encoder we use in our approach. The I/Q signal input is first passed through a 1D convolutional layer (32 kernels with size 24) to extract the spatial characteristics. The following two LSTM layers (with 128 units) are used to extract the temporal characteristics. In the end, the 1D convolutional layer (128 kernels with size 8) and the max pooling layer are used to generate the representations. 

\subsubsection{Projection Head}
The third part is a small neural network projection head $P$ that maps representations to the space where contrastive loss is applied. In our approach, we use a multilayer perceptron (MLP) with one hidden layer as the projection head, which means
\begin{equation}
    \bm{z}_i=P(\bm{r}_i)=W^{(2)}\sigma(W^{(1)}\bm{r}_i),
\end{equation}
where $\sigma$ is the ReLU activation function. It has been proved that calculating the contrastive loss on $\bm{z}_i$’s (instead of on the representations $\bm{r}_i's$ directly) improves the performance of self-supervised contrastive learning~\cite{chen2020simple}. 

\subsubsection{Contrastive Loss}
The last part is the contrastive loss function that is defined for a contrastive prediction task aiming at maximizing the agreement between examples augmented from the same signal input. We use the normalized temperature-scaled cross entropy loss (NT-Xent) \cite{sohn2016improved} as the loss function. For a given mini-batch of $M$ examples in the training process, since for each example $X$, we will generate a pair of augmented examples, there will be $2M$ data points. The two augmented versions $\hat{X}_i$ and $\hat{X}_j$ of the same input $X$ are called a positive pair. All remaining $2(M-1)$ in this batch are negative examples to them. Cosine similarity is utilized to measure the similarity between two augmented examples $\hat{X}_i$ and $\hat{X}_j$. The similarity is calculated on $\bm{z}_i$ and $\bm{z}_j$:
\begin{equation}
	\text{sim}(\bm{z}_i, \bm{z}_j)=\frac{\bm{z}_i^T\bm{z}_j}{\|\bm{z}_i\|\|\bm{z}_j\|}.
\end{equation}
Here, $\|\bm{z}_i\|$ refers to the $\ell_2$ norm of $\bm{z}_i$. The loss function for a positive pair of examples $(i, j)$ is defined as
\begin{equation}
L_{i,j} = -\log\frac{\exp(\text{sim}(\bm{z}_i,\bm{z}_j)/\tau)}{\sum_{k=1}^{2M}\mathds{1}_{[k\neq i]}\exp(\text{sim}(\bm{z}_i,\bm{z}_k)/\tau)},
\end{equation}
where $\tau$ refers to the temperature parameter of softmax and $\mathds{1}_{[k\neq i]}\in \{0,1\}$ is an indicator function equaling to 1 iff $k\neq i$.
The loss for all positive pairs $(i, j)$ is calculated and the average is used as the final contrastive loss.

\section{Evaluation} \label{sec:evaluation}
In this section, we evaluate the performance of SemiAMC using a public dataset, RadioML2016.10a \cite{o2016radio}.  We first introduce the dataset we use and then show the experimental setup, including data preprocessing and detailed implementation. Finally, we analyze the performance of SemiAMC.

\subsection{Dataset}
We use RadioML2016.10a to study the performance of SemiAMC. RadioML2016.10a is a synthetic dataset including radio signals of different modulations at varying signal-to-noise ratios (SNRs). It consists of 11 commonly used modulations (8 digital and 3 analog): WBFM, AM-DSB, AM-SSB, BPSK, CPFSK, GFSK, 4-PAM, 16-QAM, 64-QAM, QPSK, and 8PSK. For each modulation, there are 20 different SNRs from $-20$dB to $+18$dB and there are 1000 signals under each SNR. Hence, the RadioML2016.10a has $1000\times 20 \times 11=220,000$ signal examples in total. Each signal in RadioML2016.10a has 128 I/Q samples. In our approach, we put each signal in a $2\times 128$ matrix $X$.

\subsection{Experimental Setup}
We split the dataset into three parts: training, validation, and testing by a ratio of 2:1:1. Specifically, for each modulation type and SNR, we randomly divide the 1000 signals into 500 signals for training, 250 signals for validation, and 250 signals for testing. Normalized signals are used as the input. 

In the self-supervised contrastive pre-training part, we use a two-layer projection head. The first layer is a fully connected  layer  with  128  hidden  units  and  ReLU  activation. The second is a fully connected layer with 64 hidden units without an activation function. The encoder, the output of which is 128, is trained under a batch size of 512, with a total of 100 batches, and initial learning rate $1e^{-4}$ with cosine decay. In the downstream task part, we freeze the encoder and build a two-layer classifier on the representations. The first layer of the classifier is a fully connected layer with 128 neurons and ReLU activation. The second is a Softmax layer with 11 neurons, one for each modulation scheme. We apply droupout and L2 regularization to mitigate over-fitting.

We first study the performance when all the training and validation data have labels. We then study the performance when only part of the training and validation data have labels. We also study the performance of our approach when different amounts of unlabeled data are used to train the encoder. We further compute the classification accuracy separately for each SNR and modulation scheme. We run five times with different random seeds and take the average as the final performance.
  \begin{figure*}[!ht]
      \hspace{6mm}
    \begin{minipage}{0.33\linewidth}
      \includegraphics[width=55mm]{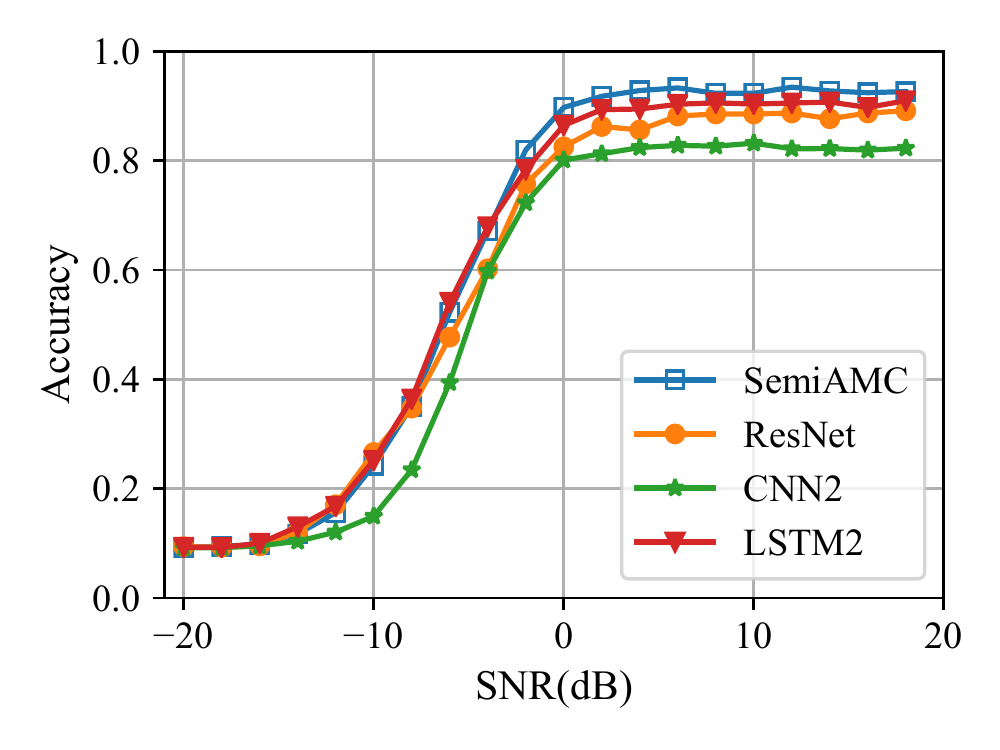}
      \vspace{-3mm}
      \caption{Comparison with other frameworks.}
      \label{fig:baseline}
    \end{minipage}
    \hfill
    \begin{minipage}{0.666\linewidth}
    \vspace{-2mm}
    \hspace{-6mm}
  \includegraphics[width=55mm]{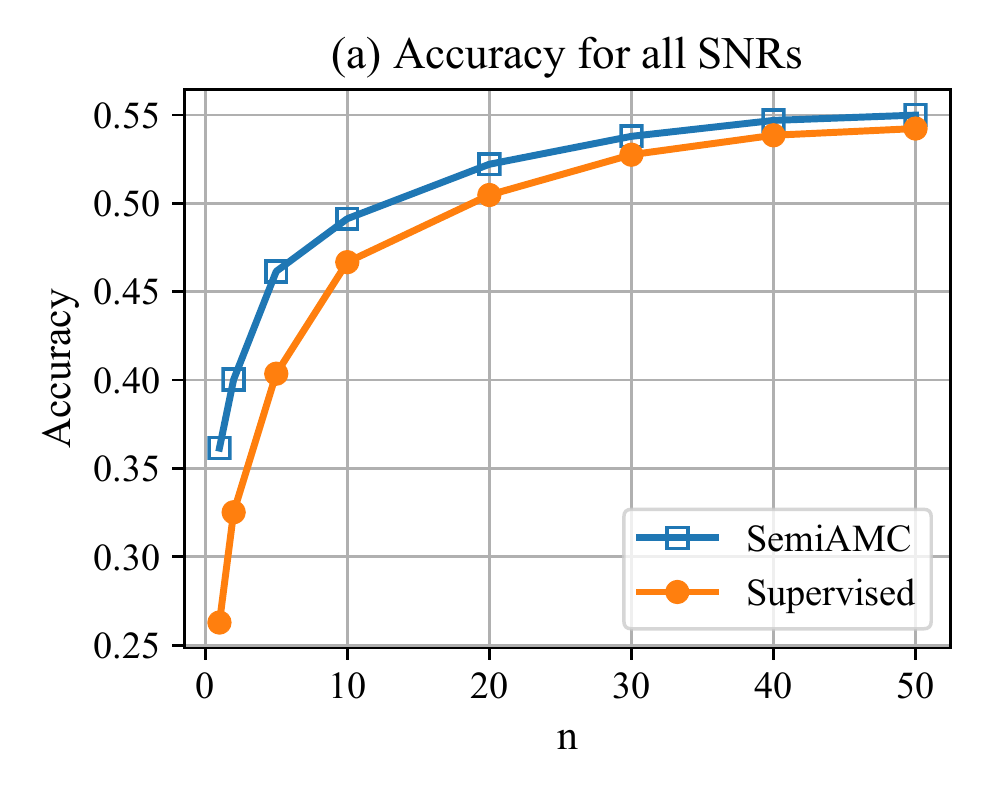}
  \includegraphics[width=55mm]{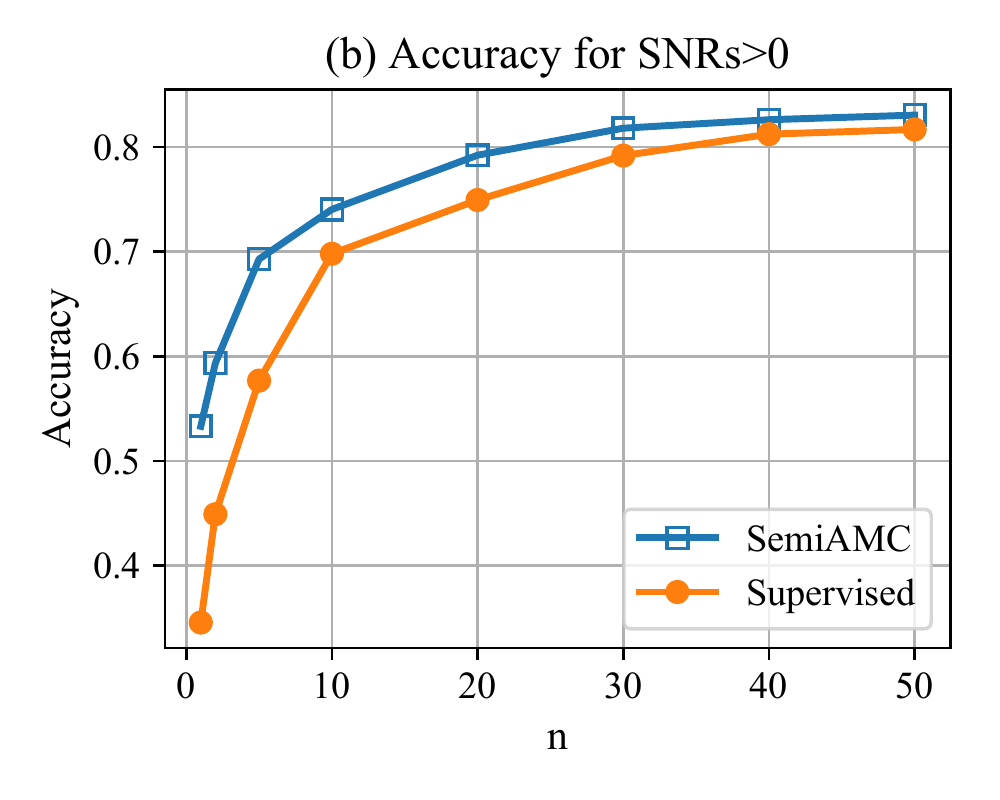}
  \vspace{-3mm}
  \caption{Accuracy under different amount of labeled data.}
  \label{fig:acc}
    \end{minipage}
  \end{figure*}

\begin{figure*}[t]
\vspace{-1mm}
  \centering
  \includegraphics[width=55mm]{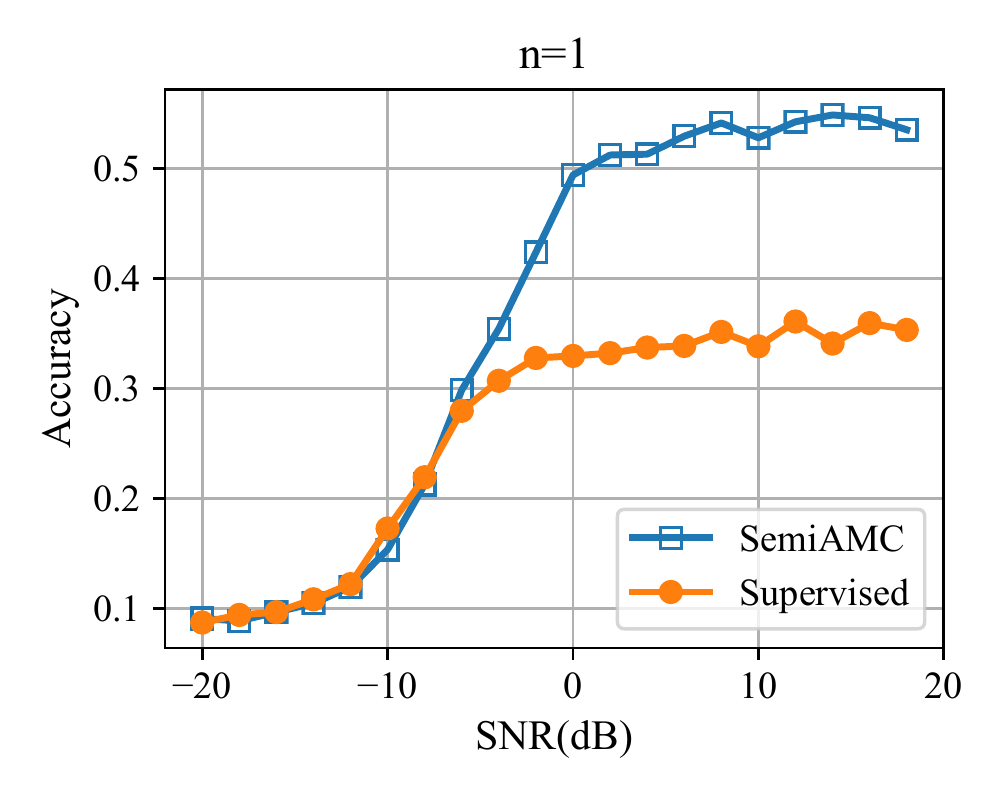}
  \includegraphics[width=55mm]{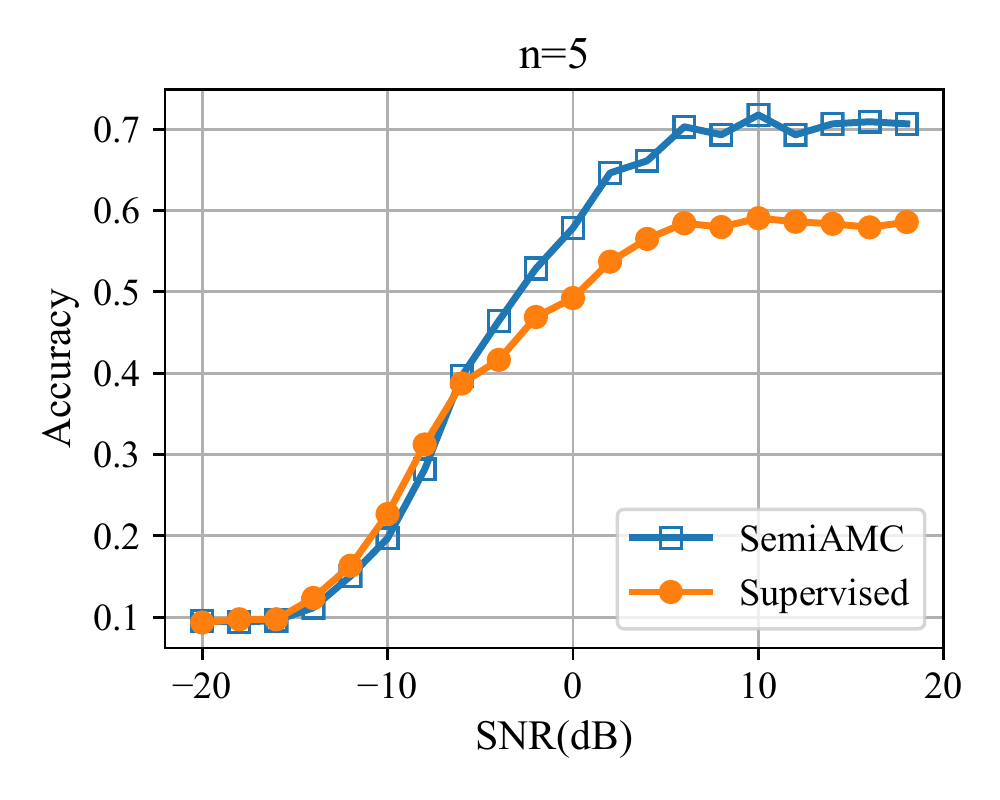}
  \includegraphics[width=55mm]{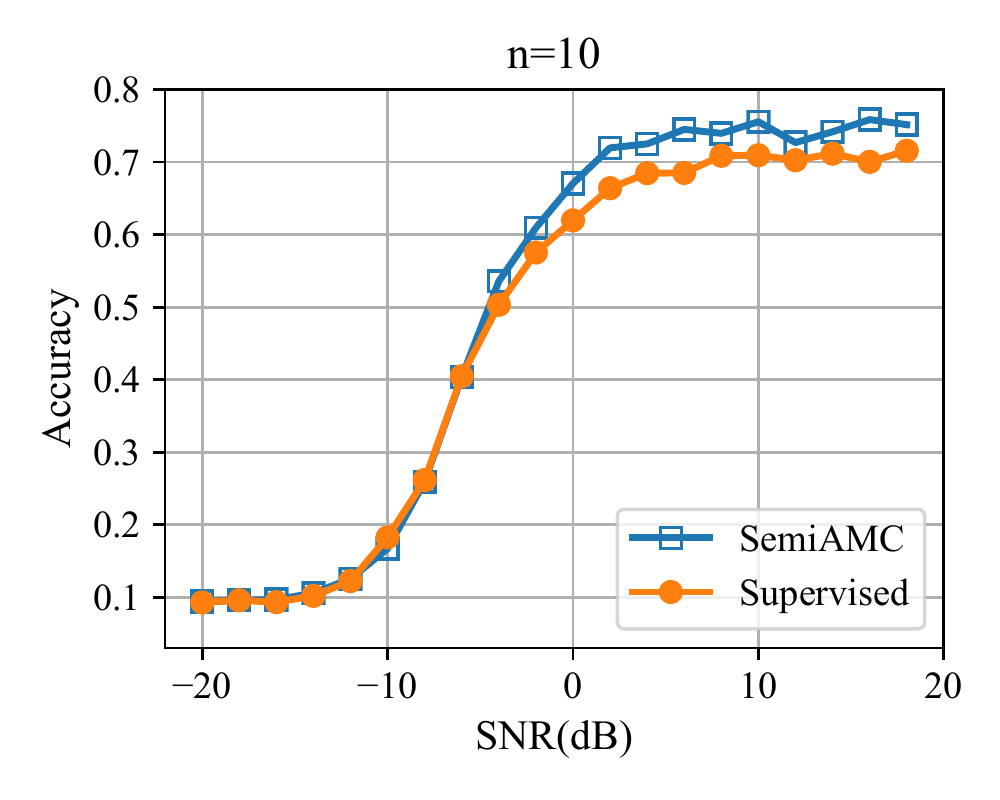}
  \vspace{-2mm}
  \caption{Accuracy for different SNRs when n=1,5,10.}
    
  \label{fig:snr}
\end{figure*}

\subsection{Comparison with Supervised Frameworks}
In this part, we compare the performance of our semi-supervised learning framework to that of previous supervised frameworks when we have enough labeled training data. Specifically, here we assume that we have labels for all the training and validation data. We first train the encoder with all signal samples in the training dataset. Then, we freeze the encoder and train the classifier based on both of the signals and labels in the training set. During this process, we also fine-tune the parameters of the encoder to get better result. We stop the training process when the validation loss does not decrease for 30 epochs and use the model with minimum validation loss to predict the classification accuracy on test set. 

We compare the performance of SemiAMC against three supervised algorithms named CNN2 \cite{o2016convolutional}, ResNet \cite{o2018over}, and LSTM2 \cite{rajendran2018deep}. As shown in Figure~\ref{fig:baseline}, SemiAMC performs the best even through there is no extra unlabeled data. The performance gain here mainly comes from the architecture of our encoder design and the data augmentation while training the encoder in self-supervised contrastive pre-training. SemiAMC clearly outperforms the supervised baselines above $-2$dB SNRs and achieves a maximum accuracy of $93.45\%$ under $12$dB SNR. 

\subsection{Performance under Different Amount of Labeled Data}
In this part, we studied the performance of SemiAMC given a large amount of unlabeled data and a small amount of labeled data. For each modulation type under each SNR, we have 500 signal samples for training, 250 signal samples for validation, and 250 signal samples for testing. We assume that for each modulation type under each SNR, only $n=1,2,5 (1\%),10,20,30,40,50 (10\%)$ signal samples have labels in the training set, and $\lceil n/2 \rceil$ signal samples have labels in the validation set. We train the encoder with all signal samples (without labels) in the training set, then we train the classifier and fine-tune the encoder with the labeled signal samples in the training set. The model with minimum validation loss is used to predict the classification accuracy on the test set. 

To study the effectiveness of the encoder learned in the self-supervised contrastive pre-training, we directly train the encoder and classifier in a supervised way with the labeled data in the training set, and then use the model with minimum validation loss to recognize the signals in the test set. We denote this approach as ``Supervised''. The classification accuracy for test signals under all SNRs and SNRS larger than zero are shown in Figure~\ref{fig:acc}. We can see that SemiAMC outperforms its corresponding supervised version. This confirms the effectiveness of the encoder trained with our self-supervised contrasive learning framework. We also observe that the gap between SemiAMC and the supervised version becomes larger with the decrease in labeled data. It suggests that our framework can effectively utilize unlabeled data. Naturally, the impact of unlabeled data is larger when less data are labeled. The performance gain of SemiAMC for SNRs larger than zero (Figure~\ref{fig:acc}(b)) are larger than those computed across all SNRs (Figure~\ref{fig:acc}(a)). It illustrates improved effectiveness of SemiAMC in dealing with higher SNR signals.

\begin{figure*}[t]
  \centering
  \includegraphics[width=44mm]{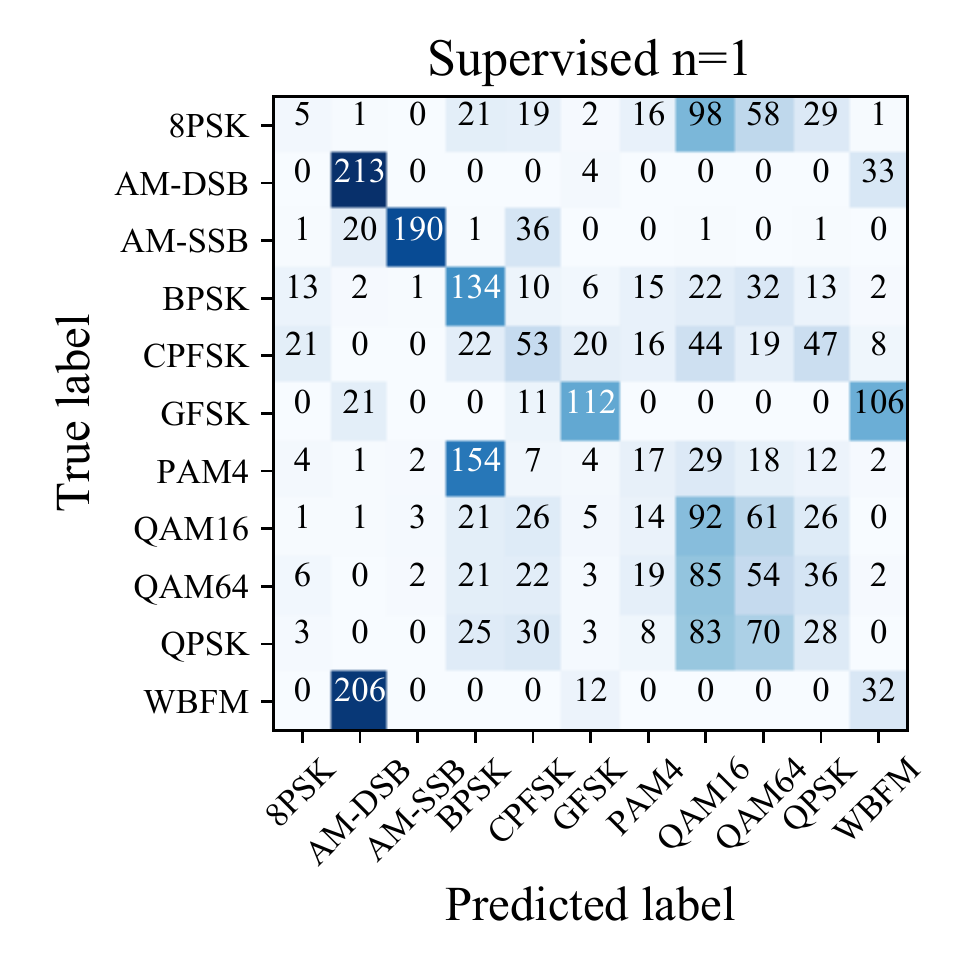}
  \includegraphics[width=44mm]{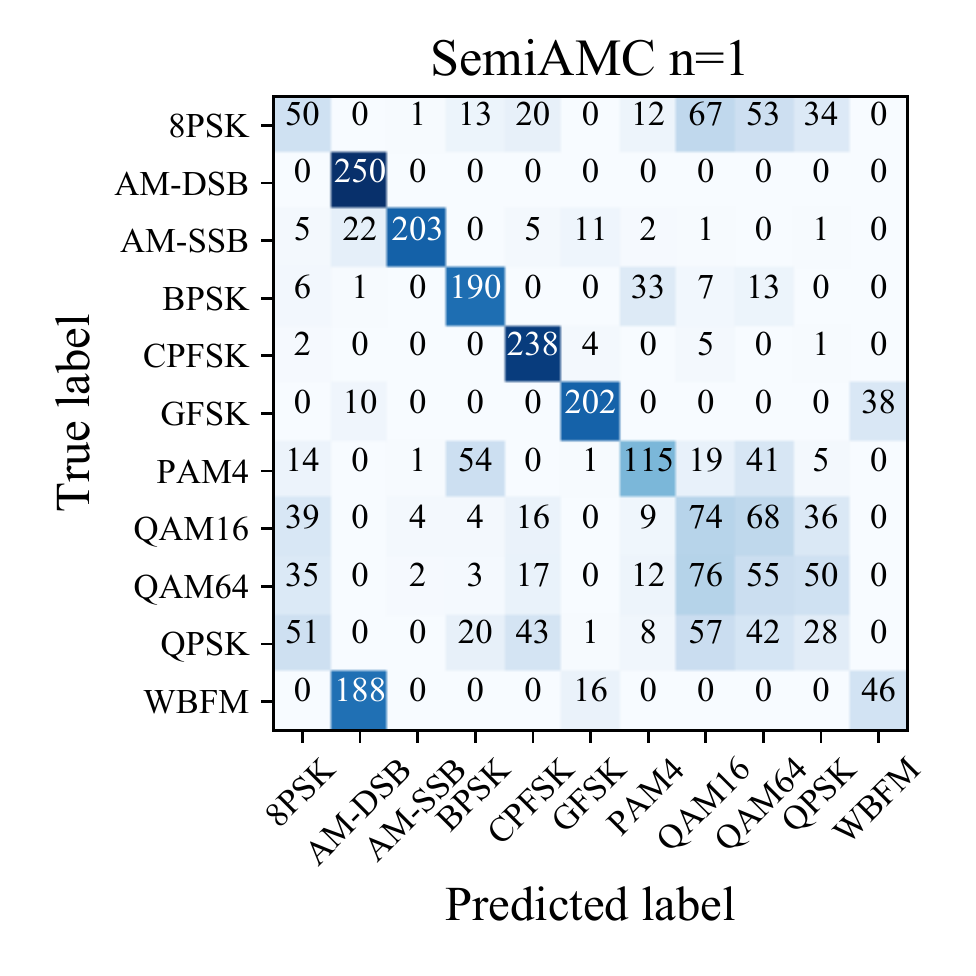}
  \includegraphics[width=44mm]{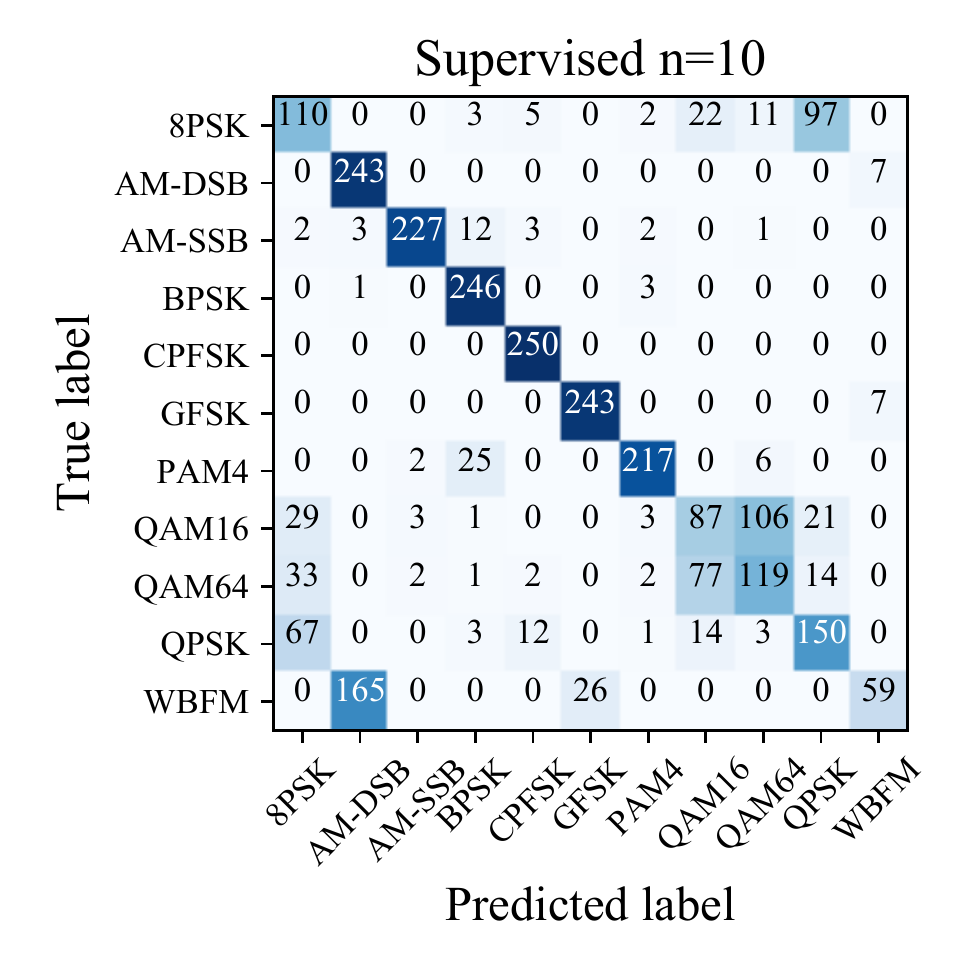}
  \includegraphics[width=44mm]{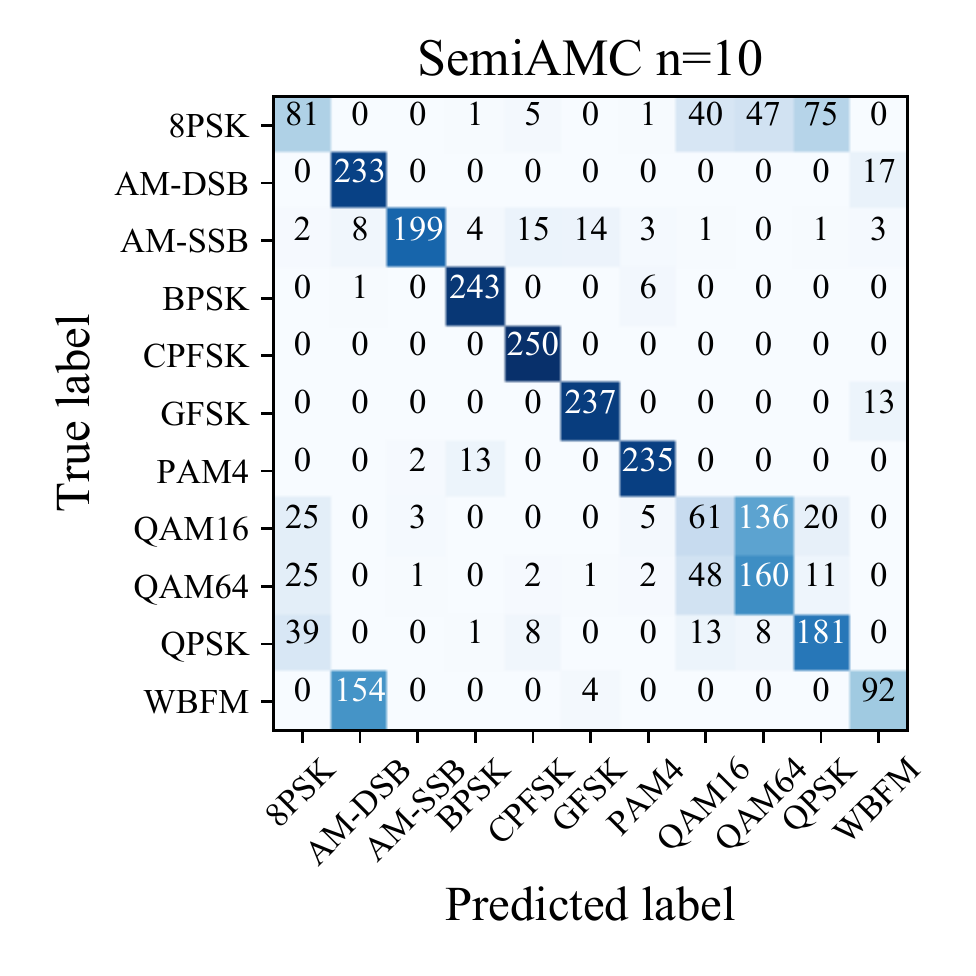}
  \caption{Confusion matrix of SemiAMC and Supervised when n=1,10.}
     \vspace{-1mm}
  \label{fig:conf}
\end{figure*}

To further understand the classification accuracy under different SNR signals when different amounts of labeled data are given, we calculate the accuracy distribution under all SNR signals when n=1,5,10. The results are shown in Figure~\ref{fig:snr}. We observe that the performance gain of SemiAMC compared with the supervised approach mainly comes from the signals with high SNR while SemiAMC performs similarly with the supervised approach when the SNRs are small. We also study the performance of SemiAMC of each modulation type when $n=1,10$ and the SNR is 10dB. We draw the corresponding confusion matrices in Figure~\ref{fig:conf}. We observe similar accuracy patterns for different modulation types between SemiAMC and its corresponding supervised approach. For example, both of them would incorrectly recognize WBFM modulated signals as AM-DSB.

\subsection{Performance under Different Amount of Unlabeled Data}
\begin{figure}[t]
  \centering
  \includegraphics[width=60mm]{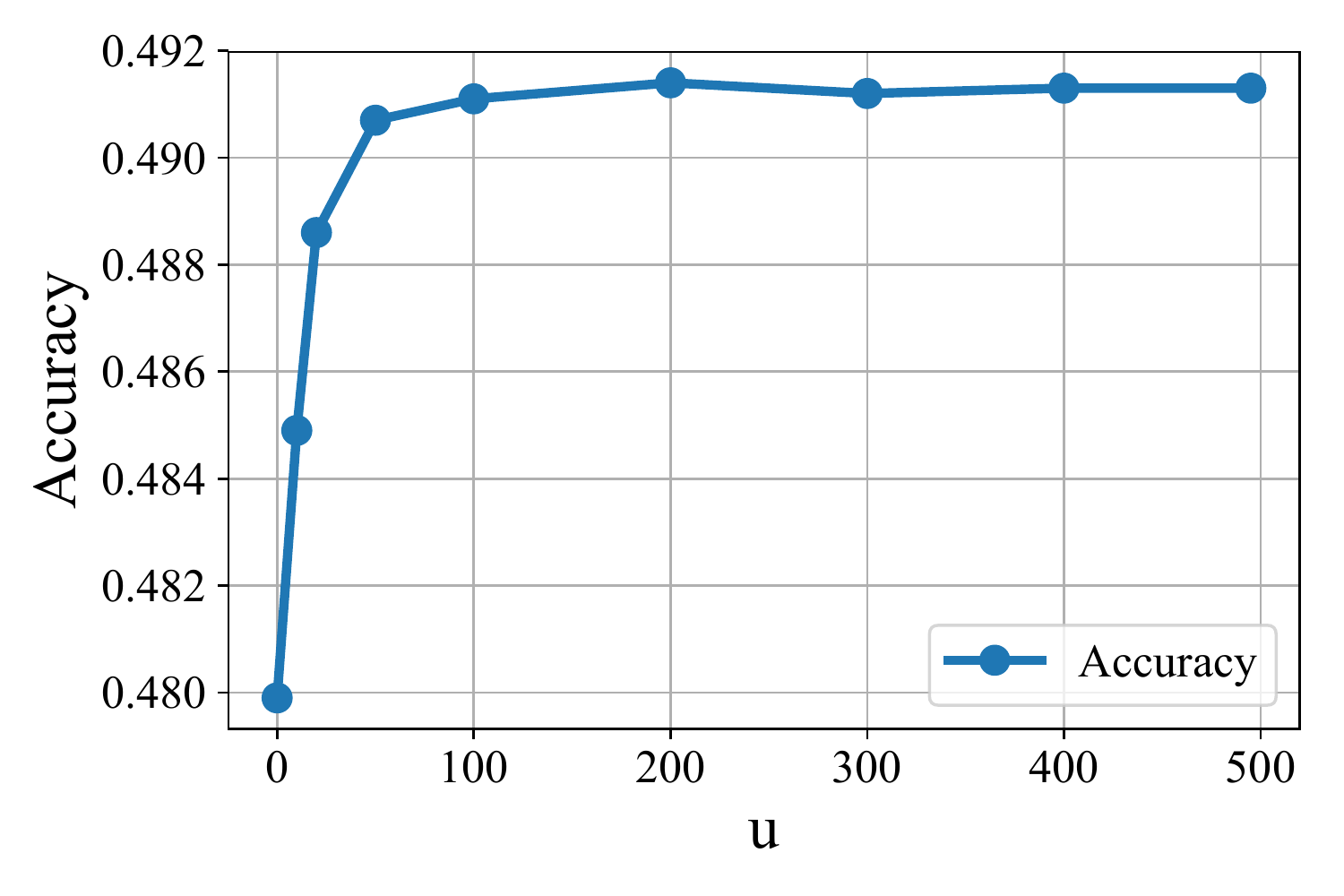}
  \vspace{-3mm}
  \caption{Accuracy under different amount unlabeled data.}
  \vspace{-2mm}
  \label{fig:unlabel}
\end{figure}

Finally, we study the recognition accuracy when different amounts of {\em unlabeled\/} data is given. Unlabeled data are used to train the encoder in the self-supervised pre-training part. In this experiment, we have 500 samples per SNR per modulation type for training. We assume that $n=10$ of the samples have labels, and use $u=0,10,20,50,100,200,300,400,490$ extra unlabeled samples besides the $n=10$ labeled samples to train SemiAMC. Figure~\ref{fig:unlabel} plots the overall accuracy versus the change in the amount of unlabeled data. It is observed that the classification accuracy increases with increasing amounts of unlabeled data (up to a certain level). 

\section{Conclusion} \label{sec:conclusion}
In this paper, we proposed a novel semi-supervised deep learning framework, called SemiAMC, to accurately recognize the modulation types of radio signals. SemiAMC efficiently utilizes unlabeled data by learning representations or features through self-supervised contrastive pre-training. We conduct several experiments on a public data set to evaluate the performance of SemiAMC. The evaluation results demonstrate a non-trivial performance gain for SemiAMC compared with supervised approaches for the same amount of labeled training data, thus verifying the effectiveness of SemiAMC at utilizing unlabeled data.

\section{Acknowledgements}
Research reported in this paper was sponsored in part by the Army Research Laboratory under Cooperative Agreement W911NF-17-2-0196 and NSF under award CPS 20-38817, and in part by The Boeing Company. The views and conclusions contained in this document are those of the authors and should not be interpreted as representing the official policies, either expressed or implied, of the Army Research Laboratory, NSF, Boeing, or the U.S. Government. The U.S. Government is authorized to reproduce and distribute reprints for Government purposes notwithstanding any copyright notation here on.

\bibliographystyle{IEEEtran}
\bibliography{MILCOM}

\end{document}